\documentclass[10pt,twocolumn,letterpaper]{article}

\usepackage{cvpr}
\usepackage{times}
\usepackage{epsfig}
\usepackage{graphicx}
\usepackage{amsmath}
\usepackage{amssymb}
\usepackage{caption}
\usepackage{xcolor}
\usepackage{contour}

\usepackage{url}

\cvprfinalcopy 

\def\cvprPaperID{11} 

\begin{document}

\title{DeepGlobe 2018: A Challenge to Parse the Earth through Satellite Images}

\author{
\.Ilke Demir$^1$, Krzysztof Koperski$^2$, David Lindenbaum$^3$, Guan Pang$^1$, \\ Jing Huang$^1$,  Saikat Basu$^1$, Forest Hughes$^1$, Devis Tuia$^4$, Ramesh Raskar$^5$ \\ \  \\ 
$^1$Facebook, $^2$DigitalGlobe, $^3$CosmiQ Works, \\ $^4$Wageningen University, $^5$ The MIT Media Lab\\
}

\twocolumn[{%
\renewcommand\twocolumn[1][]{#1}%
\maketitle
\begin{center}
    \centering
    \includegraphics[width=1\textwidth]{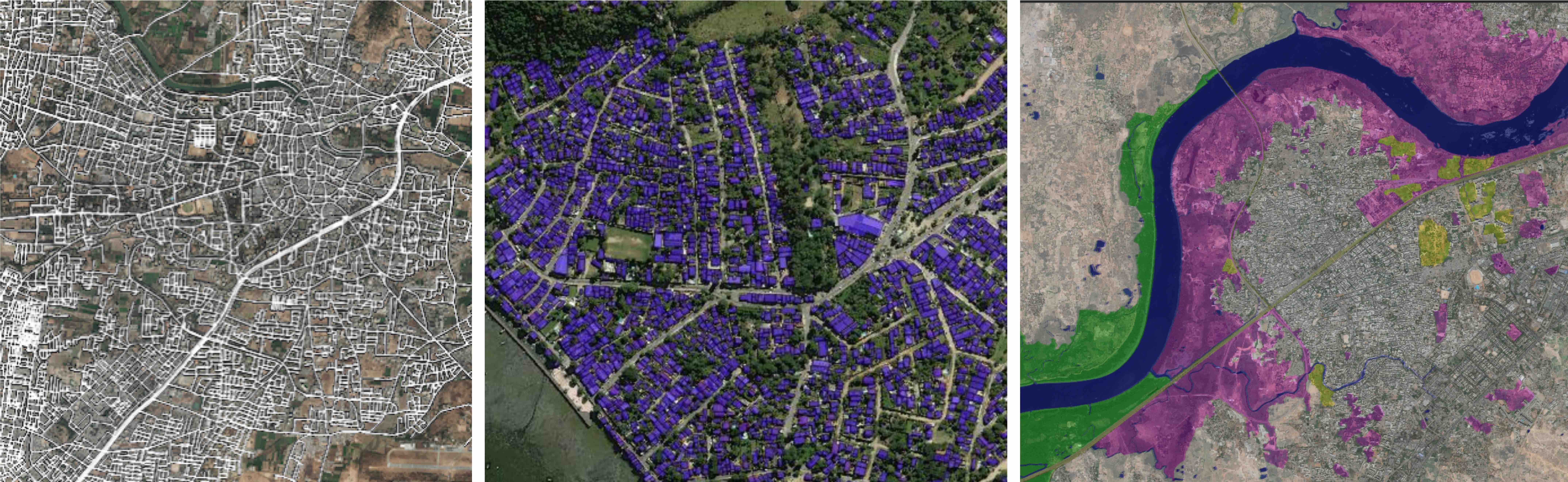}
    \captionof{figure}{\textbf{DeepGlobe Challenges:} Example road extraction, building detection, and land cover classification training images superimposed on corresponding satellite images.}\label{fig:datas}
\end{center}%
}]

\begin{abstract}
We present the DeepGlobe 2018 Satellite Image Understanding Challenge, which includes three public competitions for segmentation, detection, and classification tasks on satellite images (Figure~\ref{fig:datas}). Similar to other challenges in computer vision domain such as DAVIS\cite{davis18} and COCO\cite{coco}, DeepGlobe proposes three datasets and corresponding evaluation methodologies, coherently bundled in three competitions with a dedicated workshop co-located with CVPR 2018. 

We observed that satellite imagery is a rich and structured source of information, yet it is less investigated than everyday images by computer vision researchers. However, bridging modern computer vision with remote sensing data analysis could have critical impact to the way we understand our environment and lead to major breakthroughs in global urban planning or climate change research. Keeping such bridging objective in mind, DeepGlobe aims to bring together researchers from different domains to raise awareness of remote sensing in the computer vision community and vice-versa. We aim to improve and evaluate state-of-the-art satellite image understanding approaches, which can hopefully serve as reference benchmarks for future research in the same topic. In this paper, we analyze characteristics of each dataset, define the evaluation criteria of the competitions, and provide baselines for each task.
\end{abstract}

\section{Introduction}
As machine learning methods dominate the computer vision field, public datasets and benchmarks have started to play an important role for relative scalability and reliability of different approaches. Driven by community efforts such as ImageNet~\cite{imagenet} for object detection, COCO~\cite{coco} for image captioning, and DAVIS~\cite{davis18} for object segmentation, 
computer vision research had been able to push the limits of what we can achieve, by using the same annotated datasets and common training/validation conventions. Such datasets and corresponding challenges increase the visibility, availability, and feasibility of machine learning models, which brought up even more scalable, diverse, and accurate algorithms to be evaluated on public benchmarks. 

We observe that satellite imagery is a powerful source of information as it contains more structured and uniform data compared to traditional images. Although computer vision community has been accomplishing hard tasks on everyday image datasets using deep learning and in contrast to public datasets released for everyday media, satellite images are only recently gaining attention from the community for map composition, population analysis, effective precision agriculture, and autonomous driving tasks. 

To direct more attention to such approaches, we present DeepGlobe 2018, a Satellite Image Understanding Challenge, which (i) contains three datasets structured to solve three different satellite image understanding tasks, (ii) organizes three public challenges to propose solutions to these tasks, and (iii) gathers researchers from diverse fields to unite all expertises to solve similar tasks in a collaborative workshop. The datasets created and released for this competition may serve as (iv) fair and durable reference benchmarks for future research in satellite image analysis. Furthermore, since the challenge tasks involve ``in the wild'' forms of classic computer vision problems (e.g., image classification, detection, and semantic segmentation), these datasets have the potential to become valuable testbeds for the design of robust vision algorithms, beyond the area of remote sensing.

The three tracks for DeepGlobe are defined as follows:
\begin{itemize}

\item \textbf{Road Extraction Challenge:} In disaster zones, especially in developing countries, maps and accessibility information are crucial for crisis response. We pose the challenge of automatically extracting roads and street networks remotely from satellite images as a first step for automated crisis response and increased map coverage for connectivity.

\item \textbf{Building Detection Challenge:} As evidenced by recent catastrophic natural events, modeling population dynamics is of great importance for disaster response and recovery. Thus, modeling urban demographics is a vital task and detection of buildings and urban areas are key to achieve it. We pose the challenge of automatically detecting buildings from satellite images for gathering aggregate urban information remotely as well as for gathering detailed information about spatial distribution of urban settlements.

\item \textbf{Land Cover Classification Challenge:} Automatic categorization and segmentation of land cover is essential for sustainable development, agriculture~\cite{Alc12}, forestry~\cite{Asn05,Asn06} and urban planning~\cite{Bec17}. Therefore, we pose the challenge of classifying land types from satellite images for economic and agricultural automation solutions, among the three topics of DeepGlobe, probably as the most studied one in the intersection of remote sensing and image processing. 
\end{itemize}

We currently host three public competitions based on the tasks of extracting roads, detecting buildings, and classifying land cover types in the satellite images. 
The combined datasets include over 10K satellite images. Section~\ref{sec:data} explains the characteristics of images, details the annotation process, and introduces the division of training, validation, and test sets. Section~\ref{sec:tasks} describes the tasks in detail and proposes the evaluation metric used for each task. Section~\ref{sec:res} provides an overview of the current approaches and gives our preliminary baselines for the competitions. 

The results of the competitions will be presented in the DeepGlobe 2018 Workshop during the 2018 International Conference on Computer Vision and Pattern Recognition (CVPR) in Salt Lake City, Utah on June 18th, 2018. As of May 15st, 2018, more than 950 participants have registered in DeepGlobe competitions and there are more than 90 valid submissions in the leaderboard over the three tracks. 

\section{Datasets}\label{sec:data}
In this Section, we will discuss the dataset and imagery characteristics for each DeepGlobe track, followed by an explanation of the methodology for the annotation process to obtain the training labels.
 \subsection{Road Extraction}
There have been several datasets proposed in the literature for benchmarking algorithms for semantic segmentation of overhead imagery. Some of these can be enumerated as the TorontoCity\cite{torontocity} dataset, the ISPRS 2D semantic labeling dataset~\cite{ISPRS2}, the Mnih dataset~\cite{mnih}, the SpaceNet dataset~\cite{spacenetroad} and the ISPRS Benchmark for Multi-Platform Photogrammetry~\cite{ISPRS3}. 
  
The satellite imagery used in DeepGlobe for the road extraction challenge is sampled from the DigitalGlobe +Vivid Images dataset~\cite{vivid}. It covers images captured over Thailand, Indonesia, and India. The ground resolution of the image pixels is 50 cm/pixel. The images consist of 3 channels (Red, Green and Blue). Each of the original geotiff images are $19'584 \times 19'584$ pixels. The annotation process starts by tiling and loading these images in QGIS\cite{qgis}. Based on this tiling, we determine useful areas to sample from those countries. For designating useful areas, we sample data uniformly between rural and urban areas. After sampling we select the corresponding DigitalGlobe tiff images belonging to those areas. These images are then cropped to extract useful subregions and relevant subregions are sampled by GIS experts. (A useful subregion denotes a part of the image where we have a good relative ratio between positive and negative examples.) Also, while selecting these subregions, we try to sample interesting areas uniformly, e.g., those with different types of road surfaces (unpaved, paved, dirt roads), rural and urban areas, etc. An example of one image crop is illustrated in the left panel of Figure~\ref{fig:datas}. It is important to note that the labels generated are pixel-based, where all pixels belonging to the road are labeled, instead of labeling only the centerline. 

\begin{figure}
				\centering
				\includegraphics[width=1\linewidth]{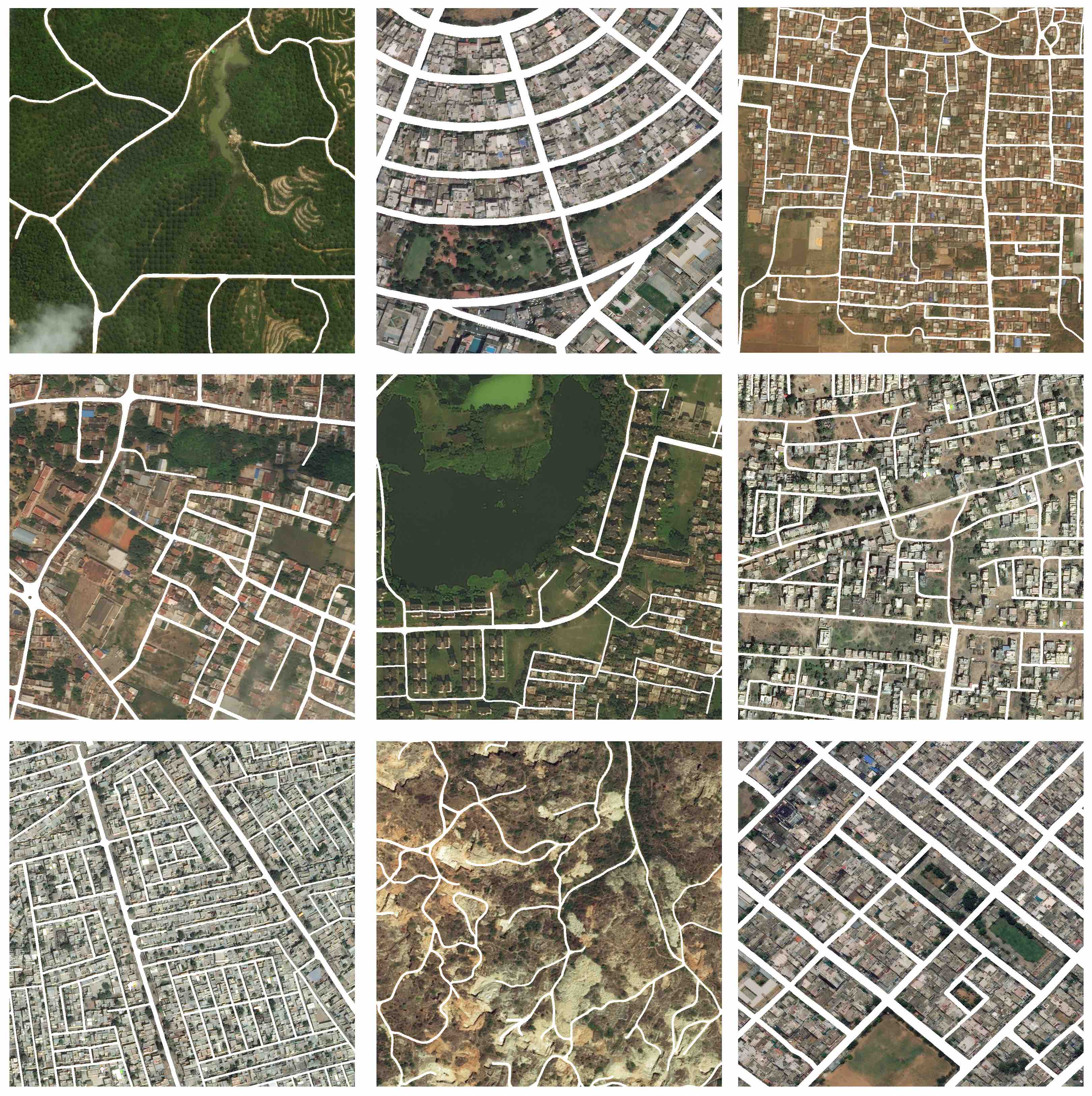}
				\caption{Road labels are annotated on top of the satellite image patches, all taken from DeepGlobe Road Extraction Challenge dataset.}
				\label{fig:roads}
			\end{figure}

The final road dataset consists of a total of $8'570$ images and spans a total land area of $2'220 km^2$. Of those, $6'226$ images (72.7\% of the dataset), spanning a total of $1'632 km^2$, were split as the training dataset. $1'243$ images, spanning $362 km^2$, were chosen as the validation dataset and $1'101$ images were chosen for testing which cover a total land area of $288 km^2$. The split of the dataset to training/validation/testing subsets is conducted by randomizing among tiles to aim for an approximate distribution of 70\%/15\%/15\%. The training dataset consists of ~4.5\% positive  and ~95.5\% negative pixels, the validation dataset consists of ~3\% positive  and ~97\% negative pixels and the test dataset consists of ~4.1\% positive  and ~95.9\% negative pixels. We selected a diverse set of patches to demonstrate road labels annotated on the original satellite images in Figure~\ref{fig:roads}. As shown, the urban morphology, the illumination conditions, the road density, and the structure of the street networks are significantly diverse among the samples.

\subsection{Building Detection}
DeepGlobe Building Detection Challenge uses the \textit{SpaceNet} Building Detection Dataset. Previous competitions on building extraction using satellite data, PRRS 2008~\cite{Aksoy2008} and ISPRS~\cite{ISPRS2, ISPRS}, were based on small areas (a few $km^2$) and in some cases used a combination of optical data and LiDAR data. The Inria Aerial Image Labeling covered $810 km^2$ area with 30cm resolution in various European and American cities \cite{maggiori2017can}. It addressed model portability between areas as some cities were included only in training data and some only in testing data. Space{N}et was the first challenge that involved large areas including cities in Asia and Africa. 

The dataset includes four areas: Las Vegas, Paris, Shanghai, and Khartoum. The labeled dataset consists of $24'586$ $200m \times 200m$ (corresponding to $650 \times 650$ pixels) non-overlapping scenes containing a total of $302'701$ building footprints across all areas. The areas are of urban and suburban nature. The source imagery is from the WorldView-3 sensor, which has both a $31cm$ single-band panchromatic and a $1.24m$ multi-spectral sensor providing 8-band multi-spectral imagery with 11-bit radiometric resolution. A GIS team at DigitalGlobe (now Radiant Solutions) fully annotated each scene, identifying and providing a polygon footprint for each building to the published specification, which were extracted to best represent the building footprint (see the central panel of Figure~\ref{fig:datas} for an example). Any partially visible rooftops were approximated to represent the shape of the building. Adjoining buildings were marked as a single building. The dataset was split 60\%/20\%/20\% for train/validation/test. As per the nature of human-based building annotation, some small errors are inevitable especially for rural areas. We leave the analysis of annotater disagreement for future work.

Each area is covered by a single satellite image, which constitutes an easier problem to solve compared to data where different parts are covered by images having different sun and satellite angles, and different atmospheric conditions. Atmospheric compensation process can process images to create data that reflects surface reflectance therefore reducing effects of atmosphere, but different shadow lengths and different satellite orientation can possibly create problems for detection algorithms if models are used to classify images acquired at different time with different sun/satellite angles.

The \textit{SpaceNet} data\cite{SpaceNetGitHub} is distributed under a Creative Commons Attribution-ShareAlike 4.0 International License and is hosted as a public dataset on Amazon Web Services and can be downloaded for free.  

\subsection{Land Cover Classification} \label{sec:lc}
Semantic segmentation started to attract more research activities as a challenging task. The ISPRS Vaihingen and Potsdam~\cite{ISPRS2} and the Zeebruges data~\cite{DFCA} are popular public datasets for this task. The ISPRS Vaihingen dataset contains 33 images of different size (on average  $2'000 \times 2'000$ pixels), with 16 fully annotated images. ISPRS Potsdam contains 38 images of size $6'000 \times 6'000$ pixels, with 24 annotated images. Annotations for both datasets have 6 classes. Vaihingen and Potsdam are both focused in urban city area, with classes limited to urban targets such as buildings, trees, cars. Zeebruges is a 7-tiles dataset (each one of size $10'000 \times 10'000$) with 8 classes (both land cover and objects), acquired both by RGB images at 5cm resolution and a LiDAR point cloud). Dstl Kaggle dataset~\cite{DsltKaggle} covered $1km^2$ of urban area with RGB and 16-band (including SWIR) WorldView-3 images. Besides urban areas, another important application of land cover classification is humanitarian studies focusing more on rural areas at mid-resolution ($\sim 30m/pixel$). For similar problems Landsat data is (i.e., crop type classification\cite{kussul17}). Still, the low resolution of Landsat data limits the information it can provide. 

The DeepGlobe Land Cover Classification Challenge is the first public dataset offering high-resolution sub-meter satellite imagery focusing on rural areas. Due to the variety of land cover types and to the density of annotations, this dataset is more challenging than existing counterparts described above. DeepGlobe Land Cover Classification Challenge dataset contains $1'146$ satellite images of size $2'448 \times 2'448$ pixels in total, split into training/validation/test sets, each with 803/171/172 images (corresponding to a split of 70\%/15\%/15\%). All images contain RGB data, with a pixel resolution of 50 cm, collected from the DigitalGlobe Vivid+ dataset as described in Section 2.1. The total area size of the dataset is equivalent to $1'716.9 km^2$.
			\begin{figure}[b]
				\centering
				\includegraphics[width=1\linewidth]{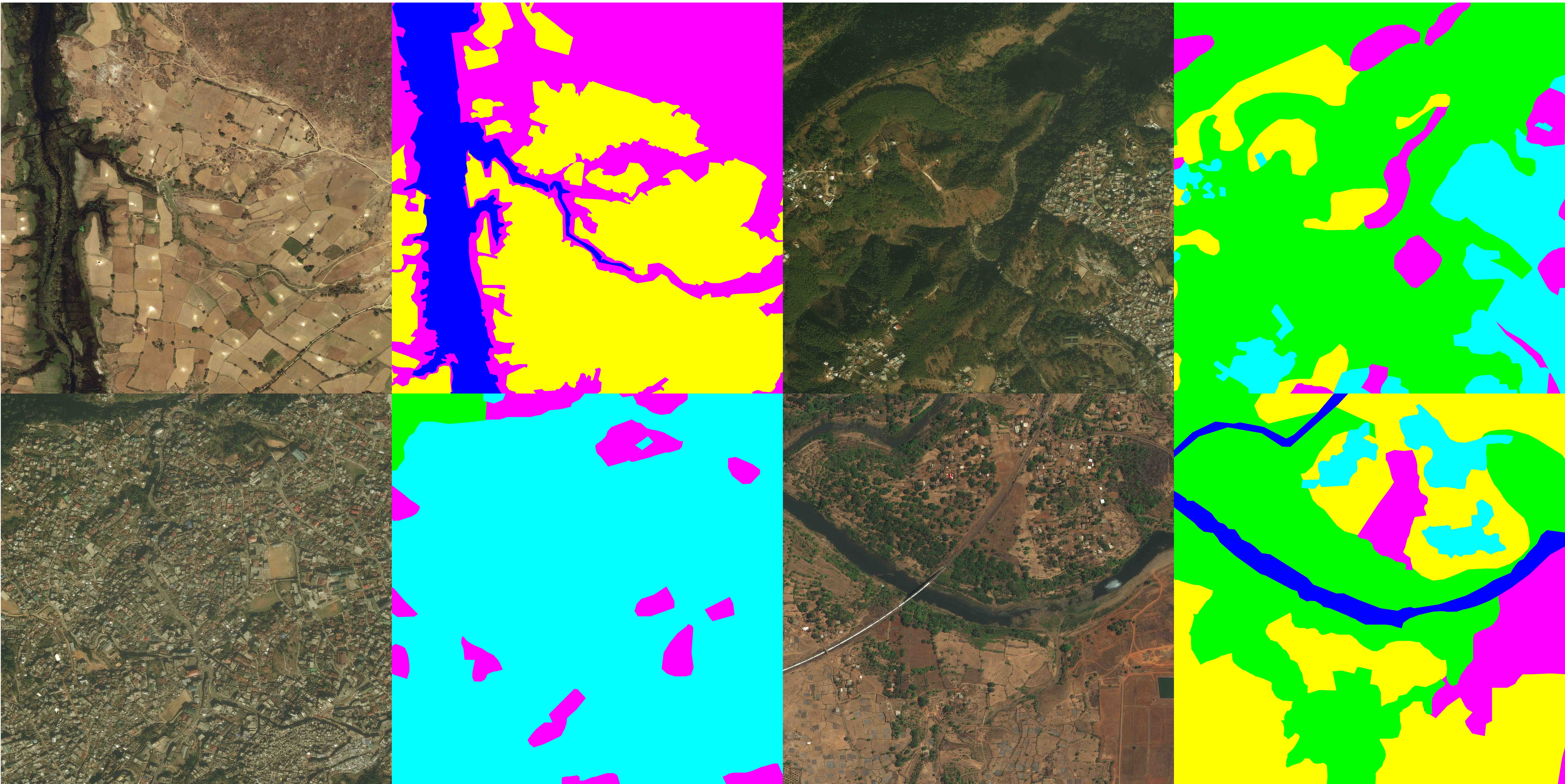}
				\caption{Some example land cover class label (right) and corresponding original image (left) pairs from interesting areas. Label colors are given in Table~\ref{tab:land_class}.}
				\label{fig:lands}
			\end{figure}


Each satellite image is paired with a mask image for land cover annotation. The mask is an RGB image with 7 classes following the Anderson Classification~\cite{LULC}. The class distributions are available in Table~\ref{tab:land_class}.

\begin{itemize}
\item Urban land: Man-made, built up areas with human artifacts.
\item Agriculture land: Farms, any planned (i.e. regular) plantation, cropland, orchards, vineyards, nurseries, and ornamental horticultural areas; confined feeding operations.
\item Rangeland: Any non-forest, non-farm, green land, grass.
\item Forest land: Any land with at least 20\% tree crown density plus clear cuts.
\item Water: Rivers, oceans, lakes, wetland, ponds.
\item Barren land: Mountain, rock, dessert, beach, land with no vegetation.
\item Unknown: Clouds and others.
\end{itemize}

\begin{table}[h]
	\centering
\begin{tabular}{ |c|c|c| } 
 \hline
 class & pixel count & proportion \\ 
 \hline
 \textcolor{cyan}{Urban} & 642.4M & 9.35\% \\ 
 \hline
 \textcolor{yellow}{Agriculture} & 3898.0M & 56.76\% \\
 \hline
 \textcolor{magenta}{Rangeland} & 701.1M & 10.21\% \\
 \hline
 \textcolor{green}{Forest} & 944.4M & 13.75\% \\
 \hline
 \textcolor{blue}{Water} & 256.9M & 3.74\% \\
 \hline
 \contourlength{0.8pt}
 \contour{black}{\textcolor{white}{Barren}} & 421.8M & 6.14\% \\
 \hline
 Unknown & 3.0M & 0.04\% \\
 \hline
\end{tabular}
	\caption{Class distributions in the DeepGlobe land cover classification dataset.}
	\label{tab:land_class}
\end{table}

The annotations are pixel-wise segmentation masks created by professional annotators (see the right hand panel in Figure~\ref{fig:datas}). The images in the dataset were sampled from full-size tiles to assure that all land cover classes have enough representation. In our specifications, we indicated that any instance of a class larger than a roughly $20m \times 20m$ would be annotated. However, land cover pixel-wise classification from high-resolution satellite imagery is still an exploratory task, and 
some small human error is inevitable. In addition, we intentionally did not annotate roads and bridges because it is already covered in the road extraction challenge. Some example labeled areas are demonstrated in Figure~\ref{fig:lands} as examples of farm, forest, and urban dominant tiles, and a mixed tile.




\section{Tasks and Evaluations}\label{sec:tasks}
In this section, we formally define the expected task in each challenge and explain the evaluation metrics used  in terms of their computation and implementation.
\subsection{Road Extraction}

We formulate the task of road extraction from satellite images as a binary classification problem. Each input is a satellite image. The solution is expected to predict a mask for the input (i.e., a binary image of the same height and width as the input with road and non-road pixel labels).

There have been previous challenges on road mask extraction, e.g., the SpaceNet \cite{spacenetroad}. Their metric was based on the Averaged Path Length Similarity (APLS) metric~\cite{APLS} that measures distance between ground truth road network represented in vector form with a solution graph also in vector form. Any proposed road graph G’ with missing edges (e.g., if an overhanging tree is inferred to sever a road) is penalized by the APLS metric, so ensuring that roads are properly connected is crucial for a high score. 

In our case, we use the pixel-wise Intersection over Union ($IoU$) score as our evaluation metric for each image, defined as Eqn. \eqref{eqn:IoU_Road}.

\begin{equation}
\label{eqn:IoU_Road}
 IoU_i = \frac{TP_i}{TP_i + FP_i + FN_i},
\end{equation}

\noindent where $TP_i$ is the number of pixels that are correctly predicted as road pixel, $FP_i$ is the number of pixels that are wrongly predicted as road pixel, and $FN_i$ is the number of pixels that are wrongly predicted as non-road pixel for image $i$. Assuming there are $n$ images, the final score is defined as the average $IoU$ among all images (Eqn.~\eqref{eqn:mIoU_Road}).

\begin{equation}
\label{eqn:mIoU_Road}
 mIoU = \frac{1}{n}\sum_{i=1}^{n}{IoU_i}
\end{equation}

 \subsection{Building Detection} 

In DeepGlobe, building detection is based on a binary segmentation task, where the input is a satellite image and the output is a list of building polygons. Multiple performance measures can be applied to score participants. PRRS 2008 \cite{Aksoy2008} challenge used 8 different performance measures. Our evaluation metric for this competition is an F1 score with the matching algorithm inspired by Algorithm 2 in the ILSVRC paper applied to the detection of building footprints~\cite{imagenet}.  This metric was decided to emphasize the importance of both accurate detection of buildings and the importance of complete identification of building outlines in an area.  Buildings with a pixel area of 20 pixels or less were discarded, as these small buildings are artifacts of the image tiling process when a tile boundary cuts a building into multiple parts. 

%
%

A detected building is scored as a true positive if the $IoU$ (Eqn.3) between the ground truth (GT) building area $A$ and the detected building area $B$ is larger than 0.5. If a proposed building intersects with multiple GT buildings, then the GT building with the largest $IoU$ value will be selected. 

\begin{equation}
	IoU=\frac{Area\left(A\cap B\right)}{Area\left(A\cup B\right)}
\end{equation}

The solution score is defined by F1 measure (Eqn. 4), where $TP$ is number of true positives, $M$ is the number of ground truth buildings and $N$ is the number of detected buildings.

\begin{equation}
	F1=2*\frac{\text{precision} * \text{recall}}{\text{precision} + \text{recall}}=\frac{2*TP}{M+N}
\end{equation}

The implementation and detailed description of scoring can be found in the SpaceNet Utilities GitHub repo \cite{SpaceNetGitHub}. 
We score each area separately and the final score is the average of scores for each area as in Eqn. 5.

\begin{equation}
	F1=\frac{F1_{AOI1}+F1_{AOI2}+F1_{AOI3}+F1_{AOI4}}{4}\\
\end{equation}

 \subsection{Land Cover Classification}

The land cover classification problem is defined as a multi-class segmentation task to detect areas of classes mentioned in Section~\ref{sec:lc}. The evaluation is computed based on the accuracy of the class labels and averages over classes are considered. The class `unknown' is removed from the evaluation, as it does not correspond to a land cover class, but rather to the presence of clouds.

Each input is a satellite image. The solution is expected to predict an RGB mask for the input, i.e., a colored image of the same height and width as the input image.
The expected result is a land cover map of same size in pixels as the input image, where the color of each pixel indicates its class label.

There have been previous challenges on road mask extraction (e.g., the TiSeLaC \cite{TiSeLaC}), which emphasized the usage of temporal information of the dataset. Our challenge, on the other hand, uses images captured at one timestamp as the input, thus more flexible in real applications. Other previous land cover / land use semantic segmentation challenges as the ISPRS~\cite{ISPRS2} or the IEEE GRSS data fusion contests~\cite{DFCA,Yok17} also used single shot ground truths and reported overall and average accuracy scores as evaluation metrics.

Similar to the road extraction challenge, we use the pixel-wise Intersection over Union ($IoU$) score as our evaluation metric. It was defined slightly differently for each class, as there are multiple categories (Eqn. \ref{eqn:IoU_LandCover}). Assuming there are $n$ images, the formulation is defined as,
\begin{equation}
\label{eqn:IoU_LandCover}
 IoU_j = \frac{\sum_{i=1}^{n}TP_{ij}}{\sum_{i=1}^{n}TP_{ij} + \sum_{i=1}^{n}FP_{ij} + \sum_{i=1}^{n}FN_{ij}},
\end{equation}

\noindent where $TP_{ij}$ is the number of pixels in image $i$ that are correctly predicted as class $j$, $FP_{ij}$ is the number of pixels in image $i$ that are wrongly predicted as class $j$, and $FN_{ij}$ is the number of pixels in image $i$ that are wrongly predicted as any class other than class $j$. Note that we have an unknown class that is not active in our evaluation (i.e., the predictions on such pixels will not be added to the calculation and thus do not affect the final score). Assuming there are $k$ land cover classes, the final score is defined as the average $IoU$ among all classes as in Eqn. \eqref{eqn:mIoU_LandCover}.

\begin{equation}
\label{eqn:mIoU_LandCover}
 mIoU = \frac{1}{k}\sum_{j=1}^{k}{IoU_j}
\end{equation}


\section{State-of-the-art and Baselines}\label{sec:res}
The tasks defined in the previous section have been explored in different datasets with different methods, some of which are also shared publicly. In this section we will introduce the state of the art approaches for each task comparing their dataset to DeepGlobe. As a baseline, we will also share our preliminary results based on current approaches on each dataset, which sets the expected success figures for the challenge participants as well as guide them to  develop novel approaches.

 \subsection{Road Extraction}
Automating the generation of road networks is extensively studied in the computer vision and computer graphics world. The procedural generation of streets~\cite{aliaga2008, chen2008} creates detailed and structurally realistic models, however written grammars are not based on the real world. On the other hand, some inverse procedural modeling (IPM) approaches~\cite{aliaga2016} process real-world data (images, LiDAR, etc.) to extract realistic representations. Following the example-based generation idea, another approach is to use already existing data resources, such as aerial images~\cite{deeproadmapper, mattyus2015}, or geostationary satellite images~\cite{ijgi, zeng2017}. Similar approaches extract road networks using neural networks for dynamic environments~\cite{wang2015} from LiDAR data~\cite{zhao2012}, using line integrals~\cite{li2016} and using image processing approaches~\cite{peteri2003, xu2014}.

\begin{figure}[t]
				\centering
				\includegraphics[width=1\linewidth]{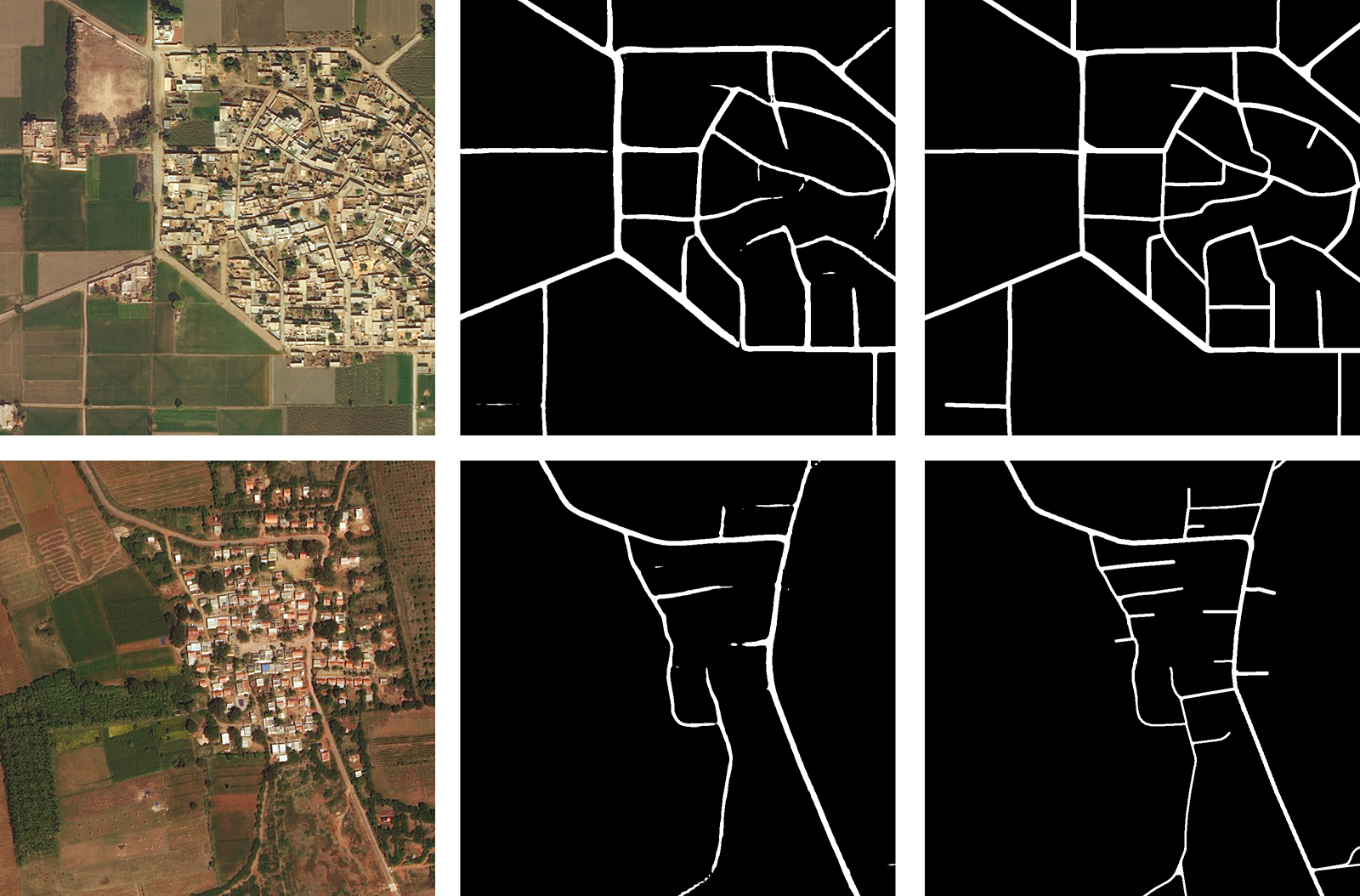}
				\caption{Example results of our road extraction method with an $IoU$ score of 0.545, with satellite image (left), extracted road mask (middle), and the ground truth (right).}
				\label{fig:road_baseline}
			\end{figure}

Similar to the experiments of~\cite{ijgi} and~\cite{deeproadmapper}, we explored our baseline approach to follow some state-of-the-art deep learning models~\cite{segnet, deeplab, resnet, unet}. In contrast to those approaches, our dataset is more diverse, spanning three countries with significant changes in topology and climate; and significantly larger in area and size. The best results were obtained by training a modified version of DeepLab~\cite{deeplab} architecture with ResNet18 backbone and Focal Loss~\cite{focalloss}. In order to provide a baseline to evaluate the network, we only added simple rotation as data augmentation, and we did not apply any post-processing to the results, only binarizing all results at a fixed threshold of 128. With this setup, we obtained an $IoU$ score of 0.545 after training 100 epochs. Two example results are given in Figure \ref{fig:road_baseline}, showing the satellite image, extracted road mask, and ground truth road mask from left to right. The vanishing roads suggest that post-processing techniques other than simple thresholding would yield more connected roads. 


\begin{table*}[t!]
    \centering
 \begin{tabular}{ | c | c | c | c | c | c | c | }
 	\hline
 	&  & Total & \multicolumn{4}{|c|}{Individual City Scores}  \\ 
 Rank	& Competitor & Score & Las Vegas & Paris & Shanghai & Khartoum \\ 
 \hline
 1	& XD\_XD  & 0.693 & 0.885 & 0.745 & 0.597 & 0.544 \\ 
 \hline
 2	& wleite & 0.643 & 0.829 & 0.679 & 0.581 & 0.483 \\ 
 \hline
 3	& nofto  & 0.579 & 0.787 & 0.584 & 0.520 & 0.424 \\
 \hline
 \end{tabular} 
	\caption{Final results from SpaceNet Building Detection Challenge Round2 (2017), as a baseline for building detection.}
	\label{tab:spacenet}
 \end {table*}

 \subsection{Building Detection} 
 
Building detection and building footprint segmentation has been subject of research for long time. Early work was based on pan-sharpened images and was using land cover classification to find
vegetation, shadows, water and man-made areas followed by segmentation and classification of segments into building/non-building areas~\cite{Aksoy2008}. Researchers sometimes transformed pixels into HSV color space,
which alleviates effects of different illumination on pitched roofs.
In~\cite{ok2013automated} the author used shadow information and vegetation/shadow/man-made classification combined with graph approach to detect buildings.

Mnih~\cite{mnih2012learning} used two locally connected NN layers followed by a fully connected layer. He also took into account omission noise (some objects are not marked in the ground truth data) by modifying loss function and mis-registration noise (such noise exists if the ground truth is not based on image, but on some other data, such as OSM~\cite{osm} or survey data) by allowing for translational noise.
Vakalopoulou et al.~\cite{vakalopoulou2015building} used convolutional layers of AlexNet to extract features that were used as input to SVM classifier which was classifying pixels into building/non-building classes.
Saito and Aoki~\cite{saito2015building} used CNN based approaches for building and road extraction. 
Liu et al.~\cite{liu2017dense} used FCN-8 segmentation network analyzing IR, R and G data with 5 convolutional layers and augmentation with a model based on nDSM (normalized Digital Surface Model) and NDVI. 
Inria competition solutions described in \cite{huang2018large} used U-Net or SegNet approaches to segmentation.
			
\begin{figure}[t]
				\centering
				\includegraphics[width=1\linewidth]{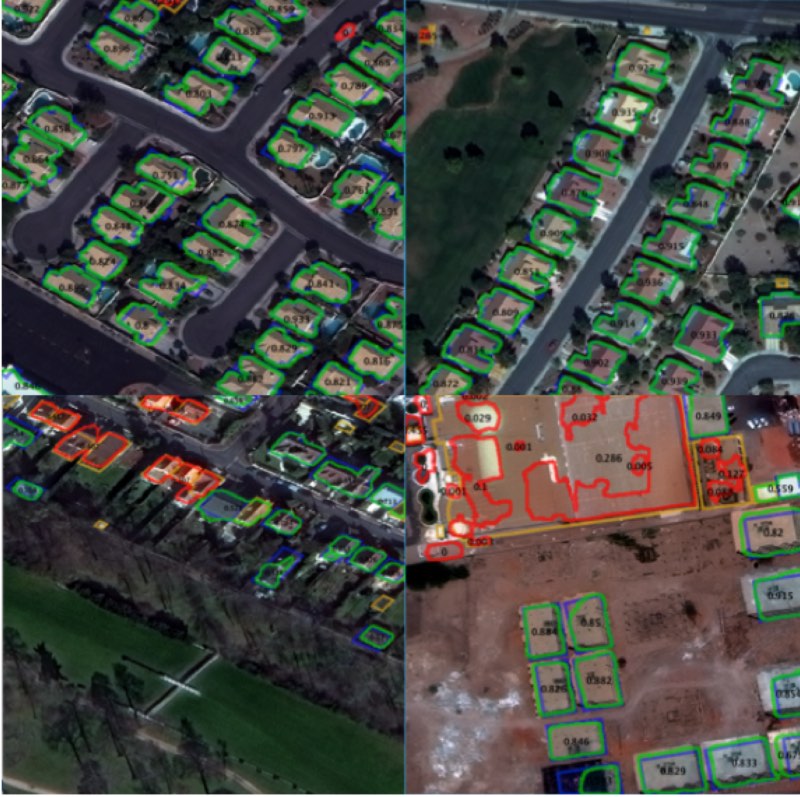}
				\caption{Results from SpaceNet baseline. From top left, clockwise: Vegas, Vegas, Khartoum, Paris. In all panels, the blue outline represents the ground truth, the green outlines are true positives, the red are false positives.}
				\label{fig:spacenet}
			\end{figure}

The current approach to building detection on our dataset uses the top scoring solutions from the SpaceNet Building Detection Round 2 result. The final results from the 2017 competition are shown in Table \ref{tab:spacenet}. It is important to note that top algorithms performed best in Las Vegas and worst in Khartoum, the visible structural organization and illumination variance in different urban morphologies are probable causes for this performance loss in the Khartoum data. The winning algorithm by competitor XD\_XD used an ensemble of 3 U-Net models~\cite{unet} to segment an 8-band multi-spectral image with additional use of OpenStreetMap~\cite{osm} data and then to extract building footprints from the segmentation output. An ensemble classifier was trained on each of the 4 AOIs individually . This segmentation approach produced high scores for $IoU$ with an average larger than 0.8, while the $IoU$ threshold for the competition is 0.5.  The algorithm struggles with small objects and in locations where buildings are very close to each other. The detailed descriptions of the algorithms can be found in \cite{SpaceNetDescription2, SpaceNetDescription1}. Figure \ref{fig:spacenet} shows the high performance of the algorithm in Las Vegas and in the bottom left you can see the algorithm has problems extracting close buildings in Paris.
			
Building detection can be followed by building footprint extraction, which can be used with DSM information to create 3D models of buildings \cite{ISPRS, arefi2013building, cg13}.
3D models can be combined with material classification and images taken from oblique angles to create accurate and realistic models for large scenes \cite{CORE3D}.

 \subsection{Land Cover Classification}
Land Cover Classification from satellite imagery is a very active research problem in remote sensing. Earlier work mostly focus on image classification, where each image is only classified to one label. Yang and Newsam\cite{yang10} used Bag-of-Visual-Words and SVM to classify a dataset of $2'100$ images containing 21 classes each with 100 images of size $256 \times 256$ pixels. The best accuracy reported\cite{yang10} was 81.19\%, and the data was released as UC Merced dataset\cite{yang10}, which later became a popular dataset for land cover image classification. Scott et al.\cite{scott17} applied DCNN-based approach on the UCM dataset and obtained a best accuracy of 98.5\% with ResNet-50.

Compared to image-level classification problem, pixel-wise classification, or semantic segmentation, started to attract more research activities as a challenging task, for which deep learning pipelines are becoming the state of the art~\cite{Zhu17}. Volpi and Tuia~\cite{volpi16} proposed to use an downsample-upsample CNN-based architecture and obtained an F1 score of 83.58\% on Vaihingen and 87.97\% on Potsdam. Audebert et al.~\cite{audebert16} trained a variant of the SegNet architecture with multi-kernel convolutional layer, and achieved 89.8\% on Vaihingen.  Marmanis and colleagues~\cite{MARMANIS2018158} achieved similar performances using an ensemble of CNNs models trained to recognized classes and using edges information between classes.  Authors in~ \cite{DFCA} performed a comparison of several state of art algorithms on the Zeebruges dataset, including kernel-based, fine-tuned VGG models and CNN trained from scratch.


We also performed pixel-wise classification on our DeepGlobe land cover data, by designing a CNN architecture based on DeepLab~\cite{deeplab} using ResNet18 backbone with atrous spatial pyramid pooling (ASPP) block and batch normalization. We used data augmentation by integrating rotations and also weighted classes based on class distributions (see Table~\ref{tab:land_class}). This approach achieved an $IoU$ score of 0.433 at epoch 30 with a 512$\times$512 patch size.

\begin{figure}
				\centering
				\includegraphics[width=1\linewidth]{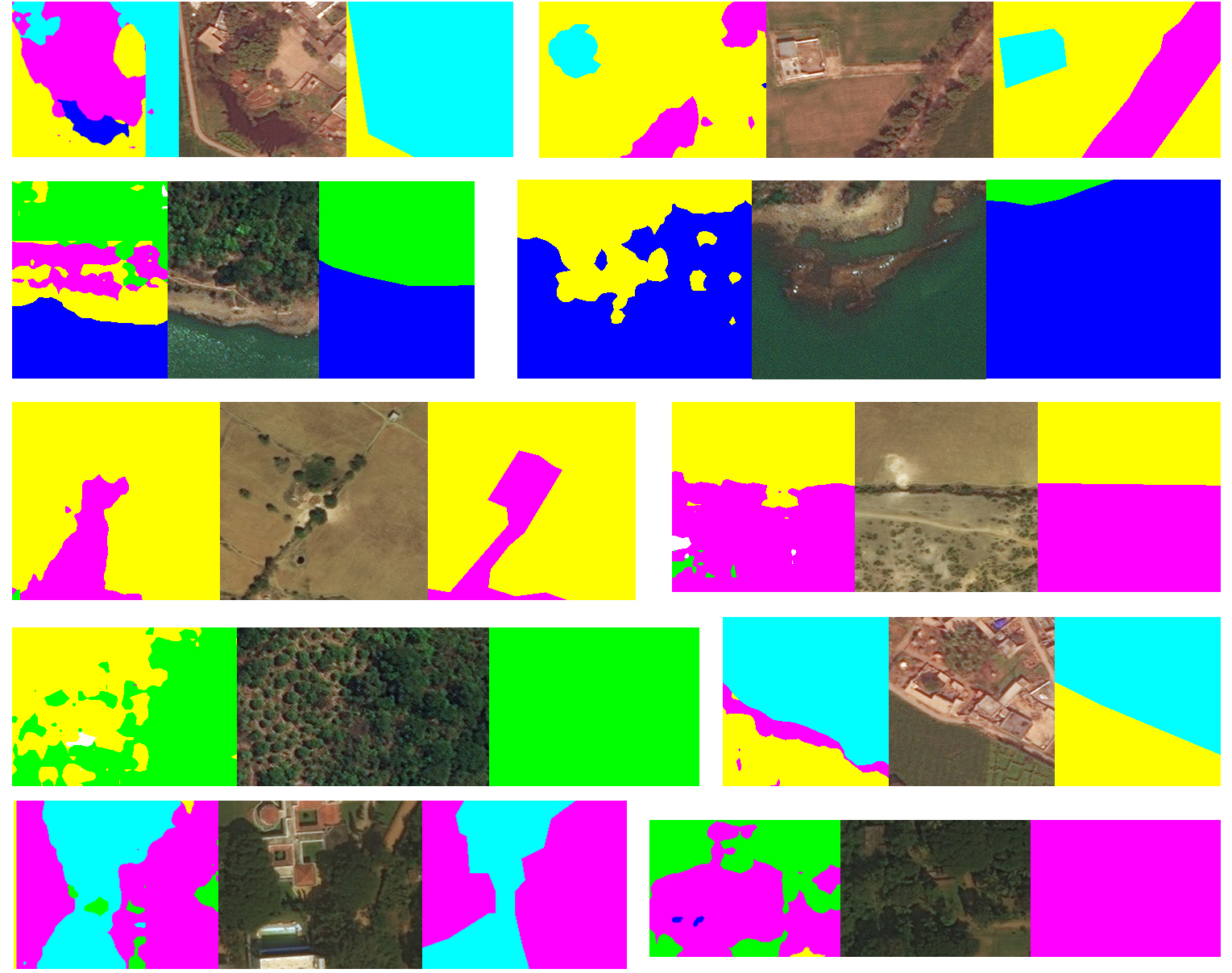}
				\caption{Some result patches (left) produced by our land cover classification baseline approach, paired with the corresponding satellite image (middle) and the ground truth (right).}
				\label{fig:land_res}
			\end{figure}

Example results are demonstrated in Figure~\ref{fig:land_res}, our result on the left, satellite image in the middle, and ground truth on the right. Note that the results and the IoU scores reported are the direct segmentation results from our model without post-processing. Harder distinctions like farms, rangelands, and forests are well-conceived by our model (third and last rows). Small structures not annotated in the ground truth, such as little ponds (top-left), and narrow beaches (second row left and fourth row right) are also correctly classified by our model. Such cases, however, decreases the $IoU$ score. Although the granularity of the segmentation looks superior to the ground truth (left of first two rows), applying a CRF or a clustering approach would improve the $IoU$ scores. 

\section{Conclusions}
We introduced the DeepGlobe Satellite Image Understanding Challenge. DeepGlobe 2018 provides datasets, challenges, and a CVPR 2018 workshop structured around the three tasks of understanding roads, buildings, and land cover types from satellite images. In this paper, we analyzed the datasets and explained the evaluation metrics for the public challenges. We also provided some baselines compared to the state-of-the-art approaches. Based on the current feedback, we believe that the DeepGlobe datasets will become valuable benchmarks in satellite image understanding, enabling more collaborative interdisciplinary research in the area, that can be fairly compared and contrasted using our benchmark, leading to new exciting developments at the intersection of computer vision, machine learning, remote sensing, and geosciences.

\section*{Acknowledgments}
We would like to thank DigitalGlobe\cite{vivid} for providing the imagery for all participants of all three challenges. We would also like to thank Facebook for sparing resources for road and land cover annotations, as well as SpaceNet\cite{SpaceNetGitHub} for sharing their building dataset with us. We would like to acknowledge the invaluable support of Prof. C. V. Jawahar and his students Suriya Singh and Anil Batra, and our colleagues Mohamed Alaoudi and Amandeep Bal for their efforts in land cover annotations.

The recognition from the community was incredible. Although we are planning to compile the results of the challenges in a follow up report, we would like to acknowledge over 950 participants, sending more than 3000 submissions to our challenge. Finally, without the DeepGlobe team, the challenge would not be as successful. It is a pleasure to work with the rest of the DeepGlobe organizing and technical committees members; namely, Daniel Aliaga, Lorenzo Torresani, Nikhil Naik, Bedrich Benes, Adam van Etten, Begum Demir, Matt Leotta, and Pau Kung.

The last word is spared to workshop sponsors for their support in DeepGlobe. Thank you Facebook, DigitalGlobe, IEEE GRSS, Uber, and CrowdAI as our gold sponsors, VSI as our silver sponsor, and Kitware as our bronze sponsor.

{\small
\bibliographystyle{ieee}
\bibliography{egbib}
}

\end{document}